\documentclass[table]{article} 
\usepackage{iclr2025_conference,times}

\usepackage{amsmath,amsfonts,bm}

\def\eqref#1{equation~\ref{#1}}

\def\1{\bm{1}}

\DeclareMathAlphabet{\mathsfit}{\encodingdefault}{\sfdefault}{m}{sl}
\SetMathAlphabet{\mathsfit}{bold}{\encodingdefault}{\sfdefault}{bx}{n}

\usepackage{hyperref}
\usepackage{url}
\usepackage{graphicx} 
\usepackage{booktabs} 
\usepackage{multirow} 
\usepackage{makecell} 
\usepackage{algorithm}
\usepackage[noend]{algpseudocode}

\usepackage{amsmath}
\usepackage{amssymb} 
\usepackage{caption}
\usepackage{wrapfig}
\usepackage{graphicx}

\usepackage{algorithmicx}

\definecolor{commentblue}{RGB}{20, 20, 200}
\definecolor{commentorange}{RGB}{255, 127, 0}

\usepackage{siunitx}    

\definecolor{highlight-green}{HTML}{E8F5E9} 
\definecolor{speedup-green}{HTML}{27AE60}

\title{Diffusion LLMs Can Do Faster-Than-AR \\Inference via Discrete Diffusion Forcing}

\author{Xu Wang$^{1}$\thanks{Equal contribution.}\,\,, 
Chenkai Xu$^{1}$\footnotemark[1]\,\,, 
Yijie Jin$^{1, 3}$, 
Jiachun Jin$^{1}$,
Hao Zhang$^{2}$,
Zhijie Deng$^{1}$\thanks{Corresponding author.} \\
{\small $^{1}$Shanghai Jiao Tong University \; $^{2}$University of California San Diego\; $^{3}$Shanghai University} \\
\texttt{\scriptsize \{wangxu60,132435xck,jiachun.jin,zhijied\}@sjtu.edu.cn, jyj2431567@shu.edu.cn}
}

\iclrfinalcopy 

\begin{document}

\maketitle


\begin{figure}[htbp]
    \centering
    \includegraphics[width=0.85\textwidth]{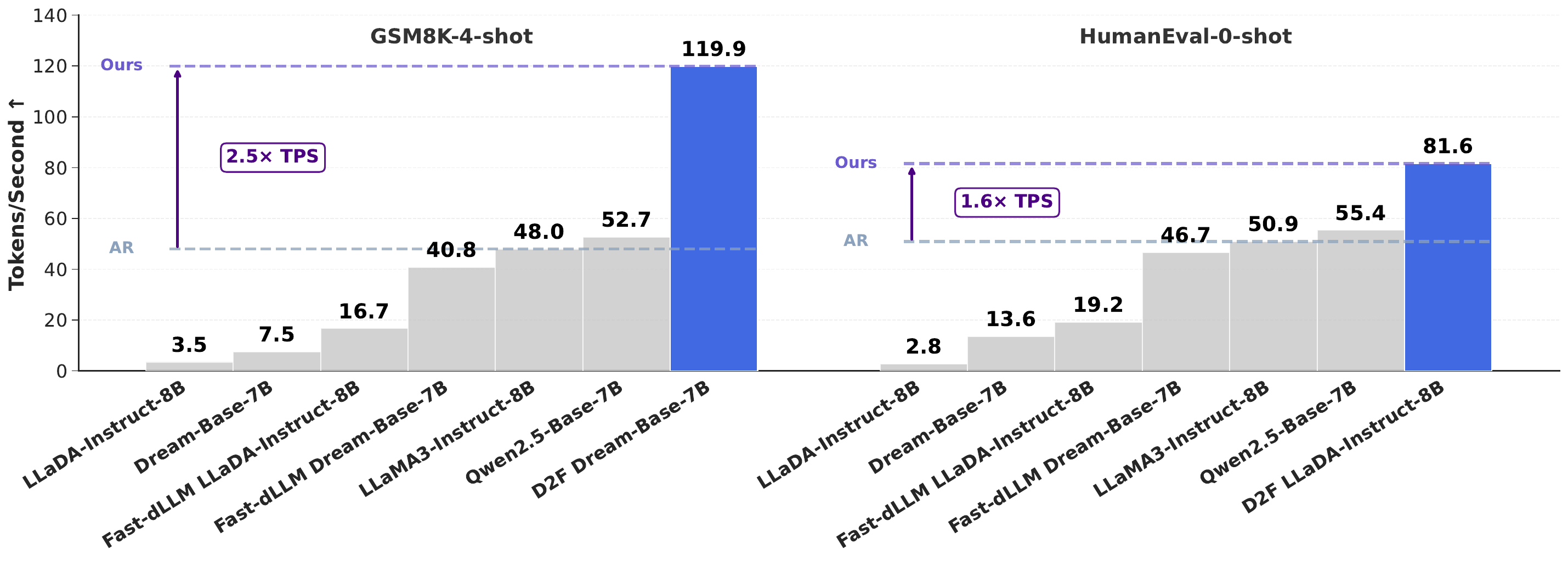}
    \captionsetup{width=0.85\textwidth, font=small}
    \caption{
    \textbf{D2F dLLMs surpass AR LLMs in inference speed for up to $2.5\times$.}
    Comparison of inference throughput among D2F dLLMs, vanilla dLLMs like Dream-Base-7B~\citep{dream2025} and LLaDA-Instruct-8B~\citep{nie2025large}, previous SOTA acceleration method Fast-dLLM~\citep{wu2025fast}, and similarly-sized AR baselines~\citep{qwen2.5, grattafiori2024llama}.
    The max generation length is set to 512.
    }
    \label{fig:inference_speed}
\end{figure}

\begin{abstract}
Diffusion Large Language Models (dLLMs) have emerged as a promising alternative to autoregressive (AR) LLMs for text generation, with the potential to decode multiple tokens in a single iteration. 
However, none of the existing open-source dLLMs have achieved superior inference speed over AR LLMs of similar size. 
This paper breaks this barrier based on a simple and effective strategy named discrete diffusion forcing (D2F). 
D2F equips dLLMs with two key capabilities: (1) block-wise autoregressive generation to enable KV cache utilization; (2) prediction of following tokens without requiring completion of prior blocks for inter-block parallel decoding. 
In this way, the vanilla dLLMs are refurbished into an AR-diffusion hybrid paradigm for efficient inference. 
D2F can be implemented with an asymmetric distillation process based on pre-trained dLLMs. 
We further propose a pipelined parallel decoding algorithm, which enables a trade-off between efficiency and efficacy. 
Empirically, 
D2F dLLMs achieve more than $\mathbf{2.5\times}$ inference speed than LLaMA3 and Qwen2.5 on GSM8K. 
Compared to vanilla dLLMs like LLaDA and Dream, the acceleration can be more than $\mathbf{50\times}$ while maintaining comparable output quality. 
The code is available at \url{https://github.com/zhijie-group/Discrete-Diffusion-Forcing}.
\end{abstract}

\begin{figure}[t]
    \centering
    \includegraphics[width=0.8\textwidth]{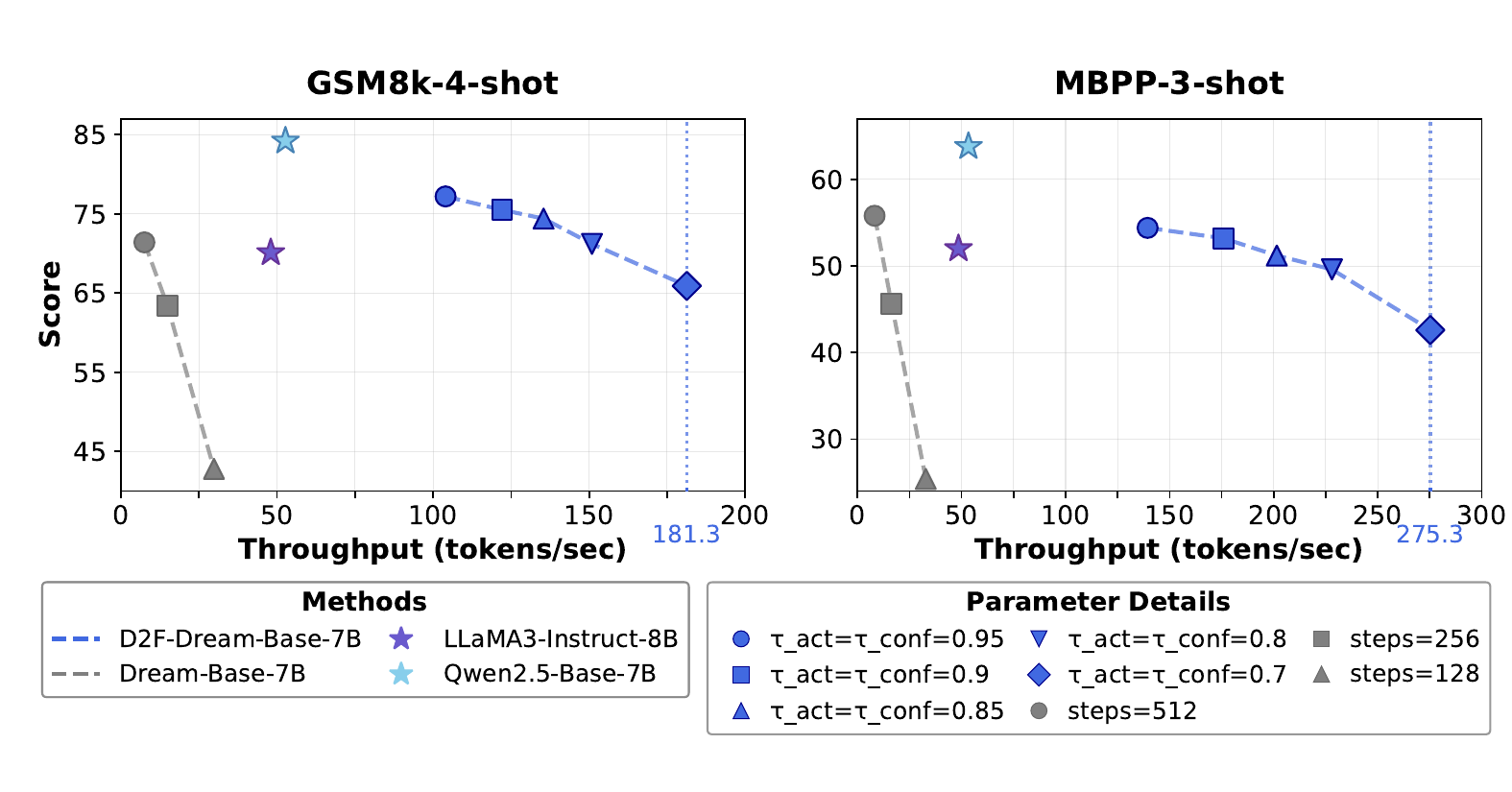}
    \captionsetup{width=\textwidth, font=small}
    \caption{
    \textbf{Throughput vs. performance trade-off.}
     As shown, D2F achieves a more favorable trade-off compared to vanilla dLLMs. 
    Refer to Section~\ref{sec:ppd} for the details of the hyperparameters $\tau_{\text{add}}$ and $\tau_{\text{conf}}$.
    }
    \label{fig:tradeoff}
\end{figure}
\section{Introduction}
Large Language Models (LLMs) have maintained a dominant position in text generation for a long time~\citep{achiam2023gpt,touvron2023llama,yang2025qwen3,touvron2023llama2,grattafiori2024llama}.
Recently, Diffusion Large Language Models (dLLMs) have emerged as a promising alternative to LLMs~\citep{dream2025,nie2025large,zhu2025llada}, 
acknowledged by their potential to generate multiple text tokens in parallel.
For example, closed-source dLLMs such as Gemini Diffusion~\citep{gemini2025} and Mercury~\citep{labs2025mercury} can yield thousands of tokens per second, 5-10 times faster than traditional autoregressive (AR) LLMs of similar size.

However, the speed merits of dLLMs have not been demonstrated within the open-source community.  
Approaches to bridge the gap include designing KV cache strategies~\citep{arriola2025block,liu2025dllm,ma2025dkv} and improving parallel sampling algorithms~\citep{wu2025fast,wei2025accelerating,hu2025accelerating}. 
For instance, Block Diffusion~\citep{arriola2025block} turns dLLMs into a block-wise sequential generation paradigm to enjoy KV cache. 
Yet, it precludes the inter-block parallelism, a crucial factor for efficient inference. 
Fast-dLLM~\citep{wu2025fast} also adopts a block-wise generation order to facilitate the reuse of states of generated tokens and incorporates confidence-based remasking for parallel decoding. 
Nonetheless, the cached states can be biased after subsequent tokens are decoded 
due to the bidirectional nature of the involved attention.

This paper achieves the first breakthrough in accelerating dLLMs to a faster-than-AR regime. 
Conceptually, we aim to embrace block-wise sequential generation to facilitate KV cache utilization, yet reject the dilemma that the decoding of subsequent blocks must wait for preceding blocks to be fully denoised. 
This implies a novel AR-diffusion hybrid paradigm, which, however, cannot be realized through naive teacher-forcing training.
This is because teacher-forcing requires complete preceding information to predict subsequent content. 
Interestingly, this insight shares the spirit with the diffusion forcing (DF)~\citep{chen2024diffusion} technique developed specifically for continuous-space diffusion models. 
In this sense, this paper forms an extension of DF to discrete data, giving rise to the discrete diffusion forcing (D2F) method for dLLM acceleration.

Concretely, D2F dLLMs learn to denoise a sequence of token blocks with monotonically increasing mask ratios in parallel. 
Naturally, preceding blocks can finish before subsequent ones, allowing their KV states to be cached for subsequent computations.
Note that we constrain the attention to be block-wise causal to ensure the KV cache remains accurate. 
For training efficiency, we distill D2F dLLMs from existing bidirectional attention dLLMs using an asymmetric distillation loss. 
In inference, we design a pipelined parallel decoding algorithm which enables inter-block parallelism and offers a decent trade-off between inference efficiency and performance (see Figure~\ref{fig:tradeoff}). 

Distilled on the Bespoke-Stratos-17k~\citep{bespoke_stratos} dataset for 12 hours with 8 NVIDIA A100-PCIe-40GB GPUs, D2F can accelerate LLaDA-Instruct-8B~\citep{nie2025large} and Dream-Base-7B~\citep{dream2025} for more than 50$\times$ without compromising average performance on mathematical and programming benchmarks,
including GSM8K~\citep{cobbe2021gsm8k}, MATH~\citep{hendrycks2021measuring}, HumanEval~\citep{chen2021codex}, and MBPP~\citep{austin2021program}. 
More importantly, D2F-Dream-Base-7B is up to 2.5$\times$ faster than LLaMA3-Instruct-8B~\citep{grattafiori2024llama} 
on GSM8K and 1.6$\times$ faster on HumanEval, establishing the first open-source dLLMs that outruns AR ones.

In summary, this work makes the following key contributions:
\begin{itemize}
    \item \textbf{D2F, a novel AR-diffusion hybrid model for text generation}, which combines the advantages of AR-style KV cache with the highly parallelized decoding of diffusion models.

    \item \textbf{The first faster-than-AR dLLMs}, which achieve up to 2.5$\times$ faster inference than prevalent AR LLMs like LLaMA3~\citep{grattafiori2024llama} and Qwen2.5~\citep{qwen2.5}. 
\end{itemize}

\section{Related Work}
\begin{figure}[t]
    \centering
    \includegraphics[width=\textwidth]{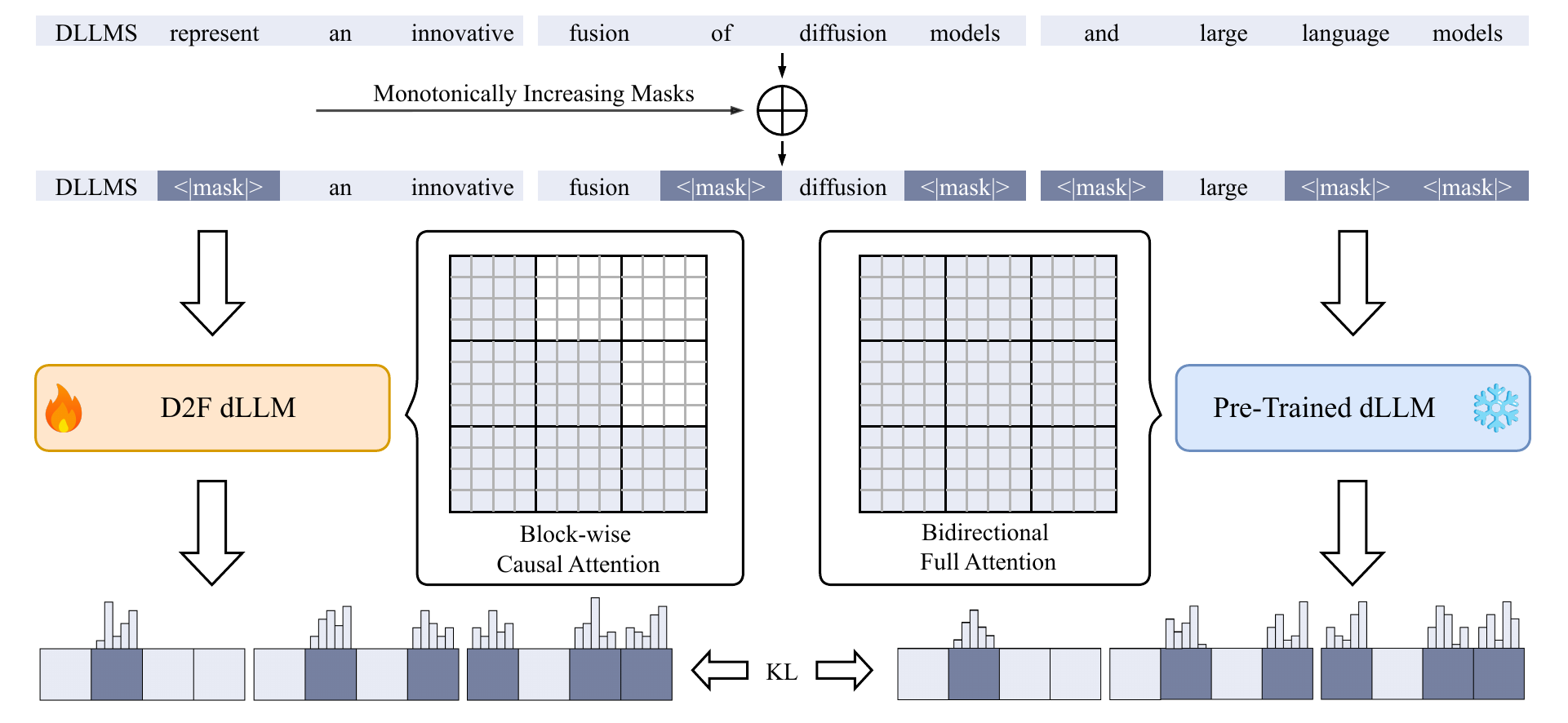}
    \captionsetup{width=\textwidth, font=small}
    \caption{
    \textbf{An overview of discrete diffusion forcing (D2F).}
    During training, the answer is divided into blocks with progressively increasing masking ratios. D2F dLLM is specified with a block-wise causal attention mask, and trained to mimic the prediction of a pre-trained bidirectional dLLM conditioned on partially denoised preceding tokens. 
    This enables inter-block parallel decoding with KV cache compatibility during inference.
    }
    \label{fig:train}
\end{figure}
\textbf{Diffusion Large Language Models (dLLMs).}
The landscape of language generation has long been defined by autoregressive models~\citep{achiam2023gpt,guo2025deepseek,liu2024deepseek}. These models, renowned for their high-quality output, are inherently limited by a sequential, token-by-token decoding process. 
To overcome this latency bottleneck, dLLMs have emerged~\citep{dream2025,nie2025large,gemini2025,labs2025mercury,zhu2025llada}. 
Instead of generating text sequentially, dLLMs operate by iteratively denoising a fully masked sequence, a process that enables the parallel prediction of all tokens at once. 
This approach, which draws from non-autoregressive principles, utilizes a bidirectional attention mechanism to achieve a more holistic understanding of context. 
Recent large-scale dLLMs, such as LLaDA~\citep{nie2025large}, trained from scratch, and Dream~\citep{dream2025}, initialized from pre-trained AR weights, have demonstrated performance competitive with leading ARMs, establishing dLLMs as a powerful alternative for high-quality, parallelizable text generation.

\textbf{Acceleration of dLLMs.}
dLLMs suffer from slower inference than autoregressive models due to incompatibility with standard KV cache and limited parallelization. Existing acceleration methods fall into two categories. First, caching-based approaches~\citep{liu2025dllm,wu2025fast,ma2025dkv} develop approximate schemes to reuse computations for static sequence parts, as standard KV cache is incompatible with bidirectional attention. Second, sampling optimization methods reduce decoding steps through confidence-aware strategies~\citep{wu2025fast}, auxiliary model guidance~\citep{israel2025accelerating}, or adaptive decoding speeds~\citep{wei2025accelerating}. However, these methods achieve limited speedups due to inherent approximations and auxiliary model overhead, often failing to match the efficiency of comparable AR~\citep{achiam2023gpt,touvron2023llama,yang2025qwen3,touvron2023llama2,grattafiori2024llama} models. Our approach fundamentally restructures generation into a block-autoregressive framework that is compatible with standard KV cache, while enabling prediction of subsequent tokens without completing previous blocks.

\textbf{AR-diffusion hybrid models.} Recent studies have explored incorporating the speed advantages of autoregressive models into diffusion-based frameworks, particularly in tasks such as video generation~\citep{yin2025slow,po2025long,sun2025ar,huang2025self}.
In video generation, it is common to model the temporal dependencies between frames using an autoregressive approach, while applying denoising within each frame independently.
This hybrid architecture leverages the acceleration benefits of KV cache enabled by autoregressive modeling, while preserving the high generation quality of diffusion-based methods.

\section{Preliminary: Diffusion Large Language Models (DLLMs)}

Diffusion models, originally developed for continuous data, have achieved state-of-the-art results in image synthesis~\citep{esser2024scaling,chen2024pixart,podell2023sdxl,labs2025flux1kontextflowmatching} 
and video generation~\citep{peng2025open,zheng2024open,yang2024cogvideox,bao2024vidu}.
In recent years, advances in the theory of discrete diffusion~\citep{austin2021structured,lou2023discrete,campbell2022continuous} have facilitated the emergence of large-scale dLLMs~\citep{nie2025large,dream2025} for text generation tasks, which have demonstrated performance competitive with their AR counterparts, while offering the potential for parallel generation.
 
The majority of successful dLLMs operate under a masked diffusion paradigm~\citep{nie2025large}. This process begins with a forward process, where an original text sequence of $L$ tokens, $Y^0 = \{y_1^0, \dots, y_L^0\}$, is progressively corrupted into a noisy sequence $Y^t$ over a continuous time schedule $t \in [0, 1]$. This corruption is routinely achieved by replacing original tokens independently with a special \texttt{[MASK]} token. 
Typically, we can define the conditional distribution as 
\begin{equation}
    q(Y^t | Y^0) = \prod_{i=1}^{L} q(y_i^t | y_i^0), \quad \text{where} \quad q(y_i^t | y_i^0) = 
    \begin{cases}
        1 - t, & \text{if } y_i^t = y_i^0 \\
        t,     & \text{if } y_i^t = \texttt{[MASK]}
    \end{cases}.
    \label{eq:forward_process}
\end{equation}
Consequently, $Y^1$ represents a fully masked sequence. 
dLLMs are designed to learn a parameterized model $p_\theta(Y^0 | Y^t)$ to reverse the forward process, hence enabling denoising from the fully masked sequence $Y^1$ to language samples $Y^0$.  
This formulation allows the model $p_\theta$ to predict all mask tokens in the sequence simultaneously at each inference step, forming the basis of dLLMs' theoretical potential for high-speed, parallel generation.

However, practical implementations of dLLMs face severe inference efficiency bottlenecks. 
First, the use of bidirectional attention mechanisms fundamentally conflicts with KV cache, leading to significant redundant computation across denoising steps. Second, the reliance on a conditional independence assumption for parallel decoding makes it hard to generate interdependent tokens~\citep{song2025ideas}, so more iterative steps are required for high-quality outputs. 
While prior works attempt to mitigate these issues, for instance, by introducing block-wise sequential generation to enable KV cache~\citep{arriola2025block} or implementing approximate KV cache~\citep{wu2025fast,liu2025dllm}, they fail to simultaneously achieve both precise KV cache and efficient parallel decoding. Consequently, no open-source dLLMs has yet matched the inference speed of AR models.

\begin{figure}[t]
    \centering
    \includegraphics[width=\textwidth]{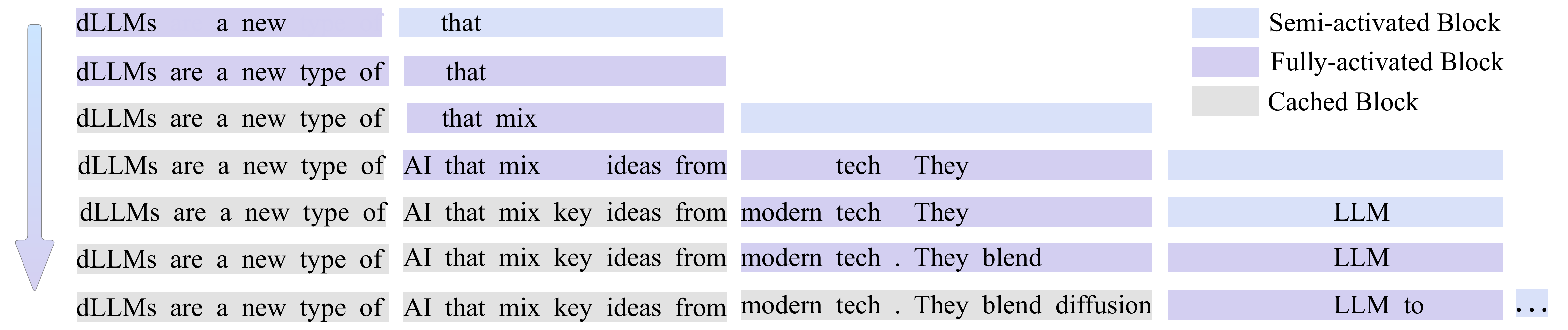}
    \captionsetup{width=\textwidth, font=small}
    \caption{
        \textbf{Visualization of the pipelined parallel decoding of D2F dLLMs.} 
        As shown, a pipeline of blocks are decoded in parallel. 
        A new block is dynamically added when the completion ratio of the last block exceeds a threshold $\tau_{add}$ ($= \frac{1}{3}$ here). 
        The new block is initially semi-activated and will transition to a fully-activated state when its predecessor reaches the completion threshold $\tau_{act}$ ($= \frac{5}{6}$ here). 
        The fully-activated blocks differ from semi-activated ones in that they would decode multiple tokens in each step more aggressively. 
    }
    \label{fig:inference}
\end{figure}

\begin{algorithm}[t]
    \caption{Asymmetric Distillation for D2F}
    \label{alg:d2f_train}
    \begin{algorithmic}[1]
        \Require Pre-trained dLLM $p_{\phi^-}$; D2F dLLM $p_{\theta}$;  block size $k$; training dataset $\mathcal{D}$.
        \While{training}
            \State Sample a sequence $Y$ from $\mathcal{D}$.
            \State Divide $Y$ into $N$ blocks $\{Y_{B_1}, \dots, Y_{B_N}\}$, each of size $k$.
            \State Sample a monotonic noise schedule $\{t_1, \dots, t_N\}$ where $t_1 < \dots < t_N$.
            \State For each $i \in \{1, \cdots, N\}$, corrupt $Y_{B_i}$ to $Y_{B_i}^{t_i}$ using Eq.~\ref{eq:forward_process}.
            \State Predict distributions for each block $Y_{B_i}^0$ with:
            \State \quad Teacher (global view): $p_{\phi^{-}}(Y_{B_i}^0 | Y_{B_1}^{t_1}, \dots, Y_{B_N}^{t_N})$
            \State \quad Student (causal view): $p_{\theta}(Y_{B_i}^0 | Y_{B_1}^{t_1}, \dots, Y_{B_i}^{t_i})$
            \State Update $p_{\theta}$ based on the loss $\mathcal{L}_\text{D2F}$ defined in Eq.~\ref{eq:d2f_loss}.
        \EndWhile
    \end{algorithmic}
\end{algorithm}

\section{Method}

The paper derives {discrete diffusion forcing (D2F)} to accelerate the inference of dLLMs to a faster-than-AR regime. 
Refer to Figures~\ref{fig:train} and~\ref{fig:inference} for an overview of the training and inference of D2F.

\subsection{Discrete Diffusion Forcing}
D2F endows dLLMs with two critical capabilities:
(1) \emph{block-level AR generation}, which enables efficient standard KV cache and significantly reduces computational redundancy; and
(2) \emph{inter-block parallel decoding}, where the model is trained to predict future blocks from incomplete, partially reconstructed predecessors to maximize the number of decoded tokens per inference step. 
This naturally gives rise to an AR-diffusion hybrid modeling paradigm, but naive teacher-forcing training, as employed in block diffusion~\citep{arriola2025block}, cannot lead to the second capability. 
To address this, D2F trains the model to perform conditional denoising of the current block based on a partially denoised prefix, enabling more coherent and efficient sequential generation.

Concretely, D2F partitions a clean sequence $Y^{0}$ into $N$ blocks of size $k$.
Let ${B_i} := \{(i-1)*k, \dots, i*k - 1\}$ denote the token indices in the $i$-th block and $Y_{B_i}$ denote the corresponding subsequence. 
In the forward process, D2F applies a monotonically increasing noise schedule ($t_1 < t_2 < \dots < t_N$) to the $N$ blocks, i.e., $Y^{t} = \{Y_{B_{1}}^{t_{1}}, \dots, Y_{B_{N}}^{t_{N}} \}$. 
Namely, the earlier blocks in $Y^{t}$ are progressively less masked (i.e., more complete), while later blocks remain increasingly masked (i.e., more uncertain). 
For the reverse process, D2F trains a $\theta$-parameterized model to characterize:
\begin{equation}
    p_\theta(Y^0 | Y^t)=\prod_{i=1}^{N} p_\theta(Y_{B_{i}}^{0}| Y_{B_{1}}^{t_{1}}, \dots, Y_{B_{i}}^{t_{i}} ). 
\end{equation}

Intuitively, 
the learned model can first finalize the decoding of preceding blocks while simultaneously advancing the denoising of subsequent ones, which effectively enables inter-block parallel decoding. 
By preserving a causal attention structure across blocks—wherein intra-block attention remains bidirectional—we can cache the KV states of already decoded blocks for exact reuse, thereby reducing redundant computations and improving inference efficiency.

\textbf{Connection to diffusion forcing~\citep{chen2024diffusion}.} 
From a high-level perspective, our approach bears a strong conceptual resemblance to diffusion forcing (DF)~\citep{chen2024diffusion}, originally developed for continuous-space diffusion models and notably applied in video generation~\citep{yin2025slow}. 
Both methods involve predicting the tokens of the next block conditioned on a noisy, incomplete premise, and our framework can be viewed as an extension of DF to discrete sequences. 
Such a principled extension motivates our naming of \emph{discrete diffusion forcing}.

\subsection{Asymmetric Distillation}
Noting that training a dLLM  with billions of parameters from scratch can be costly~\citep{nie2025large,dream2025}, we propose to distill a D2F dLLM from a pre-trained vanilla dLLM available in the open-source community. 
Let $p_{\phi^{-}}$ denote the standard bidirectional teacher dLLM and $p_\theta$ the student D2F dLLM ($\theta$ is initialized as $\phi^{-}$). 
The distillation minimizes the following loss
\begin{equation}
\label{eq:d2f_loss}
\mathcal{L}_\text{D2F} = \mathbb{E}_{ t_{1}<\dots<t_{N} } \left[ \sum_{i=1}^{N} D_{\text{KL}}\left( p_\theta(Y_{B_i}^0 | Y_{B_1}^{t_1}, \dots, Y_{B_i}^{t_i}) \Vert p_{\phi^{-}}(Y_{B_i}^0 | Y_{B_1}^{t_1},\dots, Y_{B_N}^{t_N}) \right) \right],
\end{equation}
where $D_{\text{KL}}$ represents the KL divergence aggregated over the mask tokens. 
As shown, the distillation is asymmetric---the teacher $p_{\phi^-}$ predicts for each block $Y_{B_i}^0$ with a global view of all noisy blocks while the student $p_\theta$ learns to approximate using only a causally restricted view. 
In this way, the mask prediction capabilities of existing dLLMs can be embedded into a new D2F dLLM. 
This objective also connects to CausVid~\citep{yin2025slow}, which distills existing holistic diffusion video generators into streaming ones. 
As for model architecture, the student differs from the teacher solely in attention masks---it uses the block-wise causal attention instead of a bidirectional one.

We summarize the whole algorithmic process in Algorithm~\ref{alg:d2f_train}.

\subsection{Pipelined Parallel Decoding}
\label{sec:ppd}

As illustrated in Figure~\ref{fig:inference}, we introduce a pipelined parallel decoding algorithm for the inference of D2F dLLMs. 
Specifically, we maintain a sliding window of active blocks and dynamically append a new fully masked block of tokens once the decoding progress of the last block exceeds a threshold $\tau_\text{add}$. 
The dynamic strategy significantly reduces the per-step computational cost compared to maintaining a full sequence of massive blocks throughout inference. 

Observing that aggressive decoding of a newly added block can degrade performance, we incorporate a dual-state decoding mechanism. 
Concretely, the newly added block is initialized in a semi-activated state to enable conservative parallel decoding, and will become fully activated when its predecessor has finished $\tau_\text{act}$ of the decoding---i.e., sufficient contextual information has been accumulated to support aggressive decoding of the latter block. 
Following Fast-dLLM~\citep{wu2025fast}, semi-activated blocks admit tokens with confidence above a threshold $\tau_\text{conf}$, while fully activated blocks additionally enforce the selection of the most confident token when no such token exists.

The synergy between the dynamic block management and dual-state mechanism conjoins per-step efficiency and inter-block parallelism. 
It is interesting to note that our approach also shares conceptual similarities with prior work in video generation~\citep{teng2025magi}. 
More algorithmic details are provided in Algorithm~\ref{alg:d2f_inference} and a hyperparameter analysis is in Table~\ref{tab:ablation_d2f_hyperparams_compact}.

\begin{algorithm}[t]
    \caption{Pipelined Parallel Decoding for D2F Inference}
    \label{alg:d2f_inference}
    \begin{algorithmic}[1]
        \Require D2F model $p_{\theta}$; thresholds $\tau_\text{add}, \tau_\text{act}, \tau_\text{conf}$.

        \State Initialize $Y=\{Y_{B_1}\}$ with a block of mask tokens. 
        
        \While{generation is not complete}
            \If{ {the ratio of decoded tokens in $Y_{B_{-1}}$ exceeds $\tau_\text{add}$} and {\texttt{<|EOS|>} not in $Y$}}
                \State Append a new fully masked block with semi-activated state to $Y$.
            \EndIf

            \State Forward pass of $Y$ with D2F dLLM $p_{\theta}$ using cached KV 

            \For{the active block $Y_{B_i}$ in $Y$}
                
                \State Let $J_i$ record the set of token positions in $B_i$ with $> \tau_\text{conf}$ prediction confidence
                
                \If{$Y_{B_i}$ is fully-activated and $J_i$ is $\emptyset$} 
                \State Add the token position with the highest confidence to $J_i$ 
                \EndIf

                \State Sample tokens with positions in $J_i$ and remask other positions
                
                \If{the ratio of decoded tokens in $Y_{B_{i-1}}$ exceeds $\tau_\text{act}$} \State Set $Y_{B_i}$ to be fully-activated 
                \EndIf
            \EndFor 
        \State Update KV cache for completed blocks
        \EndWhile
        
    \end{algorithmic}
\end{algorithm}

\section{Experiments}
This section details the experimental setup and presents the results of D2F dLLMs.

\begin{table*}[t]
    \centering
    
    \definecolor{brightgreen}{rgb}{0.1, 0.65, 0.2}
    
    \newcommand{\speedup}[1]{\textcolor{brightgreen}{(#1)}}

    \setlength{\tabcolsep}{3pt}

    \sisetup{detect-weight=true, detect-family=true}
    
    \begin{tabular}{
        l  
        l  
        l  
        l  
        S[table-format=3.0] 
        S[table-format=2.1] 
    }
        \toprule
        \textbf{Test Set} & \textbf{Method} & \textbf{TPS ↑} & \textbf{Latency (s) ↓} & {\textbf{Gen. Length}} & {\textbf{Score ↑}} \\
        \midrule
        
        \multirow{5}{*}{\makecell[l]{\textbf{GSM8K} \\ \small 4-shot}}
        & LLaDA-Instruct           & 7.2 \speedup{1.0x}   & 32.3 \speedup{1.0x} & 231 & 77.4          \\
        & dLLM-Cache               & 20.1 \speedup{2.8x}  & 11.5 \speedup{2.8x} & 231 & 77.5          \\
        & Fast-dLLM (Prefix-Cache) & 33.3 \speedup{4.6x}  & 7.0 \speedup{4.6x}  & 232 & 77.8          \\
        & Fast-dLLM (Dual-Cache)   & 35.2 \speedup{4.9x}  & 6.6 \speedup{4.9x}  & 232 & \textbf{78.9} \\
        \rowcolor{gray!20} 
        & \textbf{D2F-LLaDA}       & \textbf{52.5} \speedup{7.3x}  & \textbf{2.8} \speedup{11.5x} & 144 & 77.3          \\
        \midrule
        
        \multirow{5}{*}{\makecell[l]{\textbf{MBPP} \\ \small 3-shot}}
        & LLaDA-Instruct           & 0.9 \speedup{1.0x}   & 71.4 \speedup{1.0x} & 65  & \textbf{39.0} \\
        & dLLM-Cache               & 2.3 \speedup{2.6x}   & 28.3 \speedup{2.5x} & 66  & 37.0          \\
        & Fast-dLLM (Prefix-Cache) & 13.0 \speedup{14.4x} & 4.9 \speedup{14.6x} & 64  & 37.6          \\
        & Fast-dLLM (Dual-Cache)   & 15.3 \speedup{17.0x} & 3.8 \speedup{18.8x} & 58  & 36.4          \\
        \rowcolor{gray!20} 
        & \textbf{D2F-LLaDA}       & \textbf{47.6} \speedup{52.9x} & \textbf{1.4} \speedup{51.0x} & 68  & 38.0          \\
        \midrule
        
        \multirow{5}{*}{\makecell[l]{\textbf{HumanEval} \\ \small 0-shot}}
        & LLaDA-Instruct           & 2.8 \speedup{1.0x}   & 38.8 \speedup{1.0x} & 107 & 36.0          \\
        & dLLM-Cache               & 4.5 \speedup{1.6x}   & 23.3 \speedup{1.7x} & 104 & 39.0          \\
        & Fast-dLLM (Prefix-Cache) & 13.7 \speedup{4.9x}  & 7.4 \speedup{5.2x}  & 102 & 38.4          \\
        & Fast-dLLM (Dual-Cache)   & 19.2 \speedup{6.9x}  & 5.2 \speedup{7.5x}  & 100 & 35.4          \\
        \rowcolor{gray!20} 
        & \textbf{D2F-LLaDA}       & \textbf{81.6} \speedup{29.1x} & \textbf{1.6} \speedup{24.3x} & 133 & \textbf{40.2} \\
        \midrule
        
        \multirow{5}{*}{\makecell[l]{\textbf{Math} \\ \small 4-shot}}
        & LLaDA-Instruct           & 21.1 \speedup{1.0x}  & 11.5 \speedup{1.0x} & 243 & 23.7          \\
        & dLLM-Cache               & 26.9 \speedup{1.3x}  & 9.1 \speedup{1.3x}  & 246 & 23.2          \\
        & Fast-dLLM (Prefix-Cache) & 47.7 \speedup{2.3x}  & 5.2 \speedup{2.2x}  & 246 & 22.4          \\
        & Fast-dLLM (Dual-Cache)   & 42.5 \speedup{2.0x}  & 5.8 \speedup{2.0x}  & 246 & 22.4          \\
        \rowcolor{gray!20} 
        & \textbf{D2F-LLaDA}       & \textbf{90.2} \speedup{4.3x}  & \textbf{4.3} \speedup{2.7x}  & 384 & \textbf{29.1} \\
        \bottomrule
    \end{tabular}
    \caption{
   \textbf{Performance comparison of various acceleration methods on LLaDA-Instruct-8B}. 
    Speedup ratios are shown in \speedup{green}.
    All baseline methods use the default sampling configuration from the original LLaDA implementation.
    See Appendix~\ref{sec:hyperparameter} for detailed hyperparameters.
}
    \label{tab:performance_comparison_final_spaced}
\end{table*}

\subsection{Experiment Settings}

We perform evaluation on two representative dLLMs: LLaDA-Instruct-8B~\citep{nie2025large} and Dream-Base-7B~\citep{dream2025}.
For our distillation-based training, we utilize a dataset derived from the Bespoke-Stratos-17k benchmark~\citep{bespoke_stratos}. Specifically, we use two publicly available collections sourced from the HuggingFace Hub\footnote{The datasets, processed and provided by a third party, are available at: \url{https://huggingface.co/datasets/Lansechen/bs17k_collection_filtered_hard_maxlength600} and \url{https://huggingface.co/datasets/Lansechen/bs17k_collection_filtered_easy_maxlength600}.}, where a third party had previously generated responses to the problems from Bespoke-Stratos-17k using the Qwen2.5-7B model~\citep{qwen2.5}. These collections are pre-filtered to a maximum length of 600 tokens.  
The \texttt{q\_proj}, \texttt{v\_proj}, \texttt{k\_proj}, and \texttt{o\_proj} modules of D2F dLLMs are tuned with the LoRA~\citep{hu2022lora} technique, using the rank of 32, the scaling factor of 32, and the dropout rate of 0.1. 
During training, We truncate or pad all sequences to a final length of 512 tokens, and employ a block size of 16.
We applied block-wise monotonically increasing mask ratios, with a maximum threshold of 0.7 and a minimum threshold of 0.3, inspired by Block Diffusion~\citep{arriola2025block}. 
During inference, unless otherwise specified, $\tau_{\text{conf}}$ is set to 0.9, $\tau_{\text{add}}$ is set to 0.1, and $\tau_{\text{act}}$ is set to 0.95.
The models are trained for 12 hours using an AdamW optimizer with a constant learning rate of $10^{-5}$. 
All training and inference are conducted on a setup comprising 8 NVIDIA A100-PCIe-40GB GPUs.
Detailed hyperparameter configurations for inference on each benchmark are provided in the Appendix~\ref{sec:hyperparameter}.

\subsection{Main Results}

\textbf{Benchmarks.} Following established conventions, performance evaluation of D2F is conducted across mathematical reasoning and code generation benchmarks, including GSM8K~\citep{cobbe2021gsm8k}, GSM8K-CoT (Chain-of-Thought reasoning variant of GSM8K), Math~\citep{hendrycks2021measuring}, HumanEval~\citep{chen2021codex}, and MBPP~\citep{austin2021program}.

\textbf{Baselines.} Comprehensive comparisons are established between D2F and state-of-the-art acceleration strategies, including Fast-dLLM~\citep{wu2025fast} and dLLM-Cache~\citep{liu2025dllm}, implemented on LLaDA-Instruct-8B~\citep{nie2025large} and Dream-Base-7B~\citep{dream2025} architectures. Additional benchmarking against leading auto-regressive (AR) LLMs of comparable scale, specifically LLaMA3-Instruct-8B~\citep{grattafiori2024llama} and Qwen2.5-Base-7B~\citep{qwen2.5}, demonstrates the efficacy of D2F in achieving faster-than-AR inference speeds.

\textbf{Quantitative Results.}
As shown in Figure~\ref{fig:inference_speed}, D2F represents the first open-source dLLMs to surpass state-of-the-art AR LLMs in inference speed. 
The maximum generation length is set to 512 for all methods to ensure fairness. 
Concretely, D2F-Dream-Base-7B achieves a throughput of 119.9~tokens/s on GSM8K. This constitutes a 2.5$\times$ speedup over LLaMA3-Instruct-8B (48.0~tokens/s) and a 2.3$\times$ speedup over Qwen2.5-Base-7B (52.7~tokens/s).

D2F also significantly accelerates the baseline dLLMs while maintaining equivalent average performance. As detailed in Table~\ref{tab:performance_comparison_final_spaced}, D2F-LLaDA-Instruct-8B achieves a 52.9$\times$ speedup (47.6~tokens/s vs. baseline 0.9~tokens/s) with minimal performance difference (38.0 vs. baseline 39.0). Additionally, D2F-LLaDA-Instruct-8B consistently outperforms prior dLLM acceleration techniques. On MATH, it attains 90.2~tokens/s, providing a 2.1$\times$ speedup over Fast-dLLM's Dual-Cache (42.5~tokens/s). To ensure a fair comparison, the hyperparameter configuration of baseline is aligned with the paper of LLaDA~\citep{nie2025large}.

Similarly, as shown in Table~\ref{tab:acceleration_methods_comparison_optimized}, D2F-Dream-Base-7B reaches 91.2~tokens/s on GSM8K-CoT. This corresponds to a 9.6$\times$ speedup over Dream-Base-7B (9.5~tokens/s) with slight performance improvement, and a 1.8$\times$ speedup over Fast-dLLM (49.8~tokens/s). 
These results demonstrate that D2F not only surpasses existing acceleration methods but also enables dLLMs to exceed AR LLMs in throughput, substantially enhancing their practical applicability.
Since the base model struggles to generate the stop token correctly, we set a unified maximum length of 256 for all methods for a fair comparison, a setting adopted by the paper of dLLM-cache~\citep{liu2025dllm}.
More detailed hyperparameter settings are available in Appendix~\ref{sec:hyperparameter}.

\begin{table*}[t]
    \centering
    
    \definecolor{brightgreen}{rgb}{0.1, 0.65, 0.2}
    \newcommand{\speedup}[1]{\textcolor{brightgreen}{(#1)}}
    
    \setlength{\tabcolsep}{3pt}
    
    \sisetup{detect-weight=true, detect-family=true}
    
    \begin{tabular}{
        l  
        l  
        l  
        l  
        S[table-format=3.0] 
        S[table-format=2.1] 
    }
        \toprule
        \textbf{Test Set} & \textbf{Method} & \textbf{TPS ↑} & \textbf{Latency (s) ↓} & {\textbf{Gen. Length}} & {\textbf{Score ↑}} \\
        \midrule
        
        \multirow{5}{*}{\makecell[l]{\textbf{GSM8K-CoT} \\ \small 8-shot}}
        & Dream-Base               & 9.5 \speedup{1.0x}  & 26.8 \speedup{1.0x} & 255 & 75.0          \\
        & dLLM-Cache               & 26.0 \speedup{2.7x} & 9.8 \speedup{2.7x}  & 255 & 72.0          \\
        & Fast-dLLM (Prefix-Cache) & 50.3 \speedup{5.3x} & 5.1 \speedup{5.3x}  & 255 & 76.6          \\
        & Fast-dLLM (Dual-Cache)   & 49.8 \speedup{5.2x} & 5.1 \speedup{5.3x}  & 255 & 75.0          \\
        \rowcolor{gray!20}
        & \textbf{D2F-Dream}       & \textbf{91.2} \speedup{9.6x} & \textbf{2.8} \speedup{9.6x}  & 256 & \textbf{77.6} \\
        \midrule
        
        \multirow{5}{*}{\makecell[l]{\textbf{MBPP} \\ \small 3-shot}}
        & Dream-Base               & 10.4 \speedup{1.0x}  & 24.6 \speedup{1.0x} & 256 & 56.2          \\
        & dLLM-Cache               & 25.5 \speedup{2.5x}  & 10.0 \speedup{2.5x} & 256 & 52.6          \\
        & Fast-dLLM (Prefix-Cache) & 71.6 \speedup{6.9x}  & 3.6 \speedup{6.8x}  & 256 & \textbf{56.4} \\
        & Fast-dLLM (Dual-Cache)   & 73.2 \speedup{7.0x}  & 3.5 \speedup{7.0x}  & 256 & 51.0          \\
        \rowcolor{gray!20}
        & \textbf{D2F-Dream}       & \textbf{105} \speedup{10.1x} & \textbf{2.3} \speedup{10.7x} & 240 & 55.2          \\
        \midrule
        
        \multirow{5}{*}{\makecell[l]{\textbf{HumanEval} \\ \small 0-shot}}
        & Dream-Base               & 20.2 \speedup{1.0x} & 12.6 \speedup{1.0x} & 255 & 54.3          \\
        & dLLM-Cache               & 23.2 \speedup{1.1x} & 11.0 \speedup{1.1x} & 255 & \textbf{55.5} \\
        & Fast-dLLM (Prefix-Cache) & 62.4 \speedup{3.1x} & 4.1 \speedup{3.1x}  & 255 & 54.3          \\
        & Fast-dLLM (Dual-Cache)   & 60.0 \speedup{3.0x} & 4.3 \speedup{2.9x}  & 255 & 53.0          \\
        \rowcolor{gray!20}
        & \textbf{D2F-Dream}       & \textbf{73.2} \speedup{3.6x} & \textbf{3.1} \speedup{4.1x}  & 227 & 54.3          \\
        \midrule
        
        \multirow{5}{*}{\makecell[l]{\textbf{Math} \\ \small 4-shot}}
        & Dream-Base               & 9.9 \speedup{1.0x}   & 25.8 \speedup{1.0x} & 256 & 35.8          \\
        & dLLM-Cache               & 12.7 \speedup{1.3x}  & 20.2 \speedup{1.3x} & 256 & 34.5          \\
        & Fast-dLLM (Prefix-Cache) & 65.6 \speedup{6.6x}  & 3.9 \speedup{6.6x}  & 256 & \textbf{37.6} \\
        & Fast-dLLM (Dual-Cache)   & 67.0 \speedup{6.8x}  & 3.8 \speedup{6.8x}  & 256 & 37.1          \\
        \rowcolor{gray!20}
        & \textbf{D2F-Dream}       & \textbf{98.8} \speedup{10.0x} & \textbf{2.6} \speedup{9.9x}  & 256 & 35.4          \\
        \bottomrule
    \end{tabular}
    
\caption{
    \textbf{Performance comparison of various acceleration methods on Dream-Base-7B.} 
    Speedup ratios relative to the baseline are shown in \speedup{green}.
    The max generation length is set to 256.
    See Appendix~\ref{sec:hyperparameter} for detailed hyperparameters.
}
    \label{tab:acceleration_methods_comparison_optimized}
\end{table*}

\begin{figure}[t]
    \centering
    \includegraphics[width=0.9\textwidth]{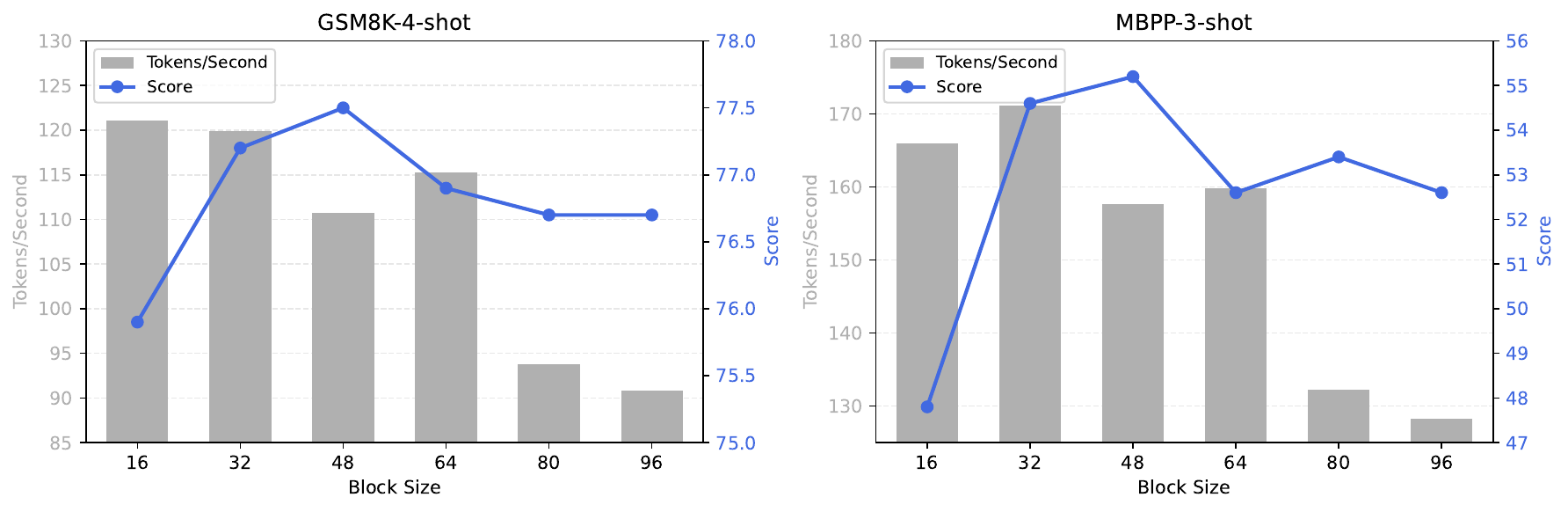}
    \captionsetup{width=\textwidth, font=small}
    \caption{
        \textbf{Ablation study on the block size during inference.}
        The block size for inference is tested with integer multiples of the training block size (16).
        All experiments were conducted with a maximum length of 512,  $\tau_{\text{conf}}=0.9$, $\tau_{\text{add}}=0.1$, and $\tau_{\text{act}}=0.95$.
    }
    \label{fig:ablation_block_size_detailed}
\end{figure}

\subsection{Ablations and Analysis}
\label{sec:ablation}
This section presents ablation studies to dissect the contributions of key components within D2F. All ablation experiments are conducted on the D2F-Dream-Base-7B model unless otherwise specified.

\textbf{Throughput-performance trade-off.}
Figure~\ref{fig:tradeoff} illustrates the throughput-performance trade-off for different methods on GSM8K and MBPP.
D2F-Dream-Base-7B employs a fixed $\tau_{\text{add}} = 0.1$ and a block size of 32, with unified thresholds $\tau_{\text{act}} = \tau_{\text{conf}}$ being varied.
Results demonstrate that D2F-Dream-Base-7B achieves significantly higher efficiency than baselines. For instance, on GSM8K, D2F-Dream-Base-7B attains 150.9~tokens/sec with a score of 71.2, achieving $3.1\times$ the throughput of LLaMA3-Instruct-8B (48.0~tokens/sec) while exceeding its score (70.1).
In contrast, Dream-Base-7B exhibits substantial performance degradation at higher throughput: reducing sampling steps from 512 to 128 causes its GSM8K score to drop from 71.4 to 42.8.
This demonstrates the superior capability of D2F-Dream-Base-7B in maintaining performance during accelerated inference.

\begin{table*}[t]
    \centering
    
    \setlength{\tabcolsep}{7pt} 
    
    \sisetup{detect-weight=true, detect-family=true, table-align-text-post=false}
    
    \begin{tabular}{
        c                   
        l                   
        S[table-format=3.1] 
        S[table-format=2.1] 
    }
        \toprule
        \textbf{$\tau_{\text{act}} = \tau_{\text{conf}}$} & \textbf{$\tau_{\text{add}}$} & {\textbf{TPS ↑}} & {\textbf{Score ↑}} \\
        \midrule
        
        \multirow{4}{*}{0.95} & 0.95 & 105.2 & 76.9 \\
                              & 0.7                 & 107.2 & 77.2 \\
                              & 0.5                 & 106.3 & 77.3 \\
                              & 0.1                 & 104.0 & \bfseries 77.7 \\
        \midrule
        
        \multirow{4}{*}{0.90} & 0.90 & 124.5 & 74.7 \\
                              & 0.7                 & 126.2 & 75.7 \\
                              & 0.5                 & 124.7 & 76.2 \\
                              & 0.1                 & 122.1 & \bfseries 76.4 \\
        \midrule
        
        \multirow{4}{*}{0.85} & 0.85 & 136.8 & 72.6 \\
                              & 0.7                 & 139.0 & 74.2 \\
                              & 0.5                 & 138.5 & 74.0 \\
                              & 0.1                 & 135.4 & \bfseries 75.0 \\
        \bottomrule
    \end{tabular}
    
\caption{
    \textbf{Ablation study of inference hyperparameter on the GSM8K-4-shot.}
    We measured the impact of $\tau_{\text{conf}}$, $\tau_{\text{act}}$, and $\tau_{\text{add}}$ on performance and speed.
    The results are tested with a block size of 32 and a maximum length of 512.
    }
    \label{tab:ablation_d2f_hyperparams_compact}
\end{table*}

\textbf{Effect of block size for inference.}
Figure~\ref{fig:ablation_block_size_detailed} demonstrates the influence of block size on inference performance and speed.
Overall, increasing block size consistently reduces throughput while yielding an initial performance improvement followed by deterioration. Optimal block size selection moderately reduces throughput but maximizes performance---for example, a block size of 48 achieves the peak GSM8K score of 77.5, surpassing smaller block sizes such as 16 with score of 75.9 despite throughput reduction.

\textbf{Ablation on hyperparameters of inference pipeline.}
Table~\ref{tab:ablation_d2f_hyperparams_compact} details an ablation study examining inference hyperparameters $\tau_{\text{conf}}$, $\tau_{\text{act}}$, and $\tau_{\text{add}}$.
We analyze the impact of the relationship between $\tau_{\text{add}}$ and $\tau_{\text{act}}$. The setting $\tau_{\text{add}} = \tau_{\text{act}}$ represents a single-state pipeline where new blocks are always fully-activated upon addition.
In contrast, our dual-state approach ($\tau_{\text{add}} < \tau_{\text{act}}$) allows for a more conservative initial state.
The results demonstrate that this dual-state strategy outperforms the single-state baseline.
For instance, with $\tau_{\text{act}}=0.85$, $\tau_{\text{add}}=0.7$ increases the score from 72.6 to 74.2 while improving throughput from 136.8 to 139.0 TPS.
Further reducing $\tau_{\text{add}}$ yields additional score improvements at the cost of marginal throughput reduction.

\textbf{Advantages of D2F training over fully random masks.}
As shown in Table~\ref{tab:ablation_noise_schedule_combined}, the D2F training strategy significantly outperforms a baseline utilizing random noise schedules per block. On MBPP, D2F demonstrates comprehensive superiority, achieving a 5.0-point score improvement (54.6 versus 49.6) and 24.0 TPS gain at $\tau_{\text{act}}=0.95$. 
These results substantiate that the structured progressive noising of D2F delivers a more effective training objective than fully random masking.

\begin{table}[t]
    \centering

    \setlength{\tabcolsep}{7pt} 
    \sisetup{
        detect-weight=true, 
        detect-family=true, 
        table-align-text-post=false
    }
    
    \begin{tabular}{
        l                   
        c                   
        l                   
        S[table-format=3.1] 
        S[table-format=2.1] 
    }
        \toprule
        \textbf{Benchmark} & \textbf{$\tau_{\text{act}}$} & \textbf{Model} & {\textbf{TPS ↑}} & {\textbf{Score ↑}} \\
        \midrule
        
        \multirow{6}{*}{\shortstack{MBPP \\ 3-shot}} 
                              & \multirow{2}{*}{0.95} & random & 147.2 & 49.6 \\
                              &                       & D2F & \bfseries 171.2 & \bfseries 54.6 \\
        \cmidrule(l){2-5} 
                              & \multirow{2}{*}{0.90} & random & 148.5 & 48.6 \\
                              &                       & D2F & \bfseries 175.5 & \bfseries 53.2 \\
        \cmidrule(l){2-5}
                              & \multirow{2}{*}{0.85} & random & 150.6 & 47.6 \\
                              &                       & D2F & \bfseries 177.5 & \bfseries 52.6 \\
        \midrule
        
        \multirow{6}{*}{\shortstack{GSM8k \\ 4-shot}} 
                              & \multirow{2}{*}{0.95} & random & 114.5 & 77.1 \\
                              &                       & D2F & \bfseries 119.9 & \bfseries 77.2 \\
        \cmidrule(l){2-5}
                              & \multirow{2}{*}{0.90} & random & 116.7 & \bfseries 76.5 \\
                              &                       & D2F & \bfseries 123.5 & 76.4 \\
        \cmidrule(l){2-5}
                              & \multirow{2}{*}{0.85} & random & 118.8 & \bfseries 75.9 \\
                              &                       & D2F & \bfseries 124.8 & 75.4 \\
        \bottomrule
    \end{tabular}
    
    \caption{
        \textbf{Ablation study on the noise scheduling strategy on the MBPP-3-shot and GSM8k-4-shot.} We compare our D2F method against a random baseline, which is trained with an independent random noise schedule for each block. All experiments use a maximum length of 512, a block size of 32, and fixed hyperparameters $\tau_{conf}=0.9$ and $\tau_{add}=0.1$. 
    }
    \label{tab:ablation_noise_schedule_combined}
\end{table}

\section{Conclusion}
In this work, we introduce Discrete Diffusion Forcing (D2F), a novel training paradigm for dLLMs.
D2F employs a generation scheme that conditions on partially predicted tokens from previous blocks to predict the next block, thereby supporting KV cache and enabling parallel generation across multiple blocks, resulting in significantly faster inference.
Empirically, extensive experiments demonstrate that D2F achieves the milestone of being the first dLLM to support faster-than-AR inference.


\bibliography{iclr2025_conference}

\begin{thebibliography}{44}
\providecommand{\natexlab}[1]{#1}
\providecommand{\url}[1]{\texttt{#1}}
\expandafter\ifx\csname urlstyle\endcsname\relax
  \providecommand{\doi}[1]{doi: #1}\else
  \providecommand{\doi}{doi: \begingroup \urlstyle{rm}\Url}\fi

\bibitem[Achiam et~al.(2023)Achiam, Adler, Agarwal, Ahmad, Akkaya, Aleman, Almeida, Altenschmidt, Altman, Anadkat, et~al.]{achiam2023gpt}
Josh Achiam, Steven Adler, Sandhini Agarwal, Lama Ahmad, Ilge Akkaya, Florencia~Leoni Aleman, Diogo Almeida, Janko Altenschmidt, Sam Altman, Shyamal Anadkat, et~al.
\newblock Gpt-4 technical report.
\newblock \emph{arXiv preprint arXiv:2303.08774}, 2023.

\bibitem[Arriola et~al.(2025)Arriola, Gokaslan, Chiu, Yang, Qi, Han, Sahoo, and Kuleshov]{arriola2025block}
Marianne Arriola, Aaron Gokaslan, Justin~T Chiu, Zhihan Yang, Zhixuan Qi, Jiaqi Han, Subham~Sekhar Sahoo, and Volodymyr Kuleshov.
\newblock Block diffusion: Interpolating between autoregressive and diffusion language models.
\newblock \emph{arXiv preprint arXiv:2503.09573}, 2025.

\bibitem[Austin et~al.(2021{\natexlab{a}})Austin, Johnson, Ho, Tarlow, and Van Den~Berg]{austin2021structured}
Jacob Austin, Daniel~D Johnson, Jonathan Ho, Daniel Tarlow, and Rianne Van Den~Berg.
\newblock Structured denoising diffusion models in discrete state-spaces.
\newblock \emph{Advances in neural information processing systems}, 34:\penalty0 17981--17993, 2021{\natexlab{a}}.

\bibitem[Austin et~al.(2021{\natexlab{b}})Austin, Odena, Nye, Bosma, Michalewski, Dohan, Jiang, Cai, Terry, Le, et~al.]{austin2021program}
Jacob Austin, Augustus Odena, Maxwell Nye, Maarten Bosma, Henryk Michalewski, David Dohan, Ellen Jiang, Carrie Cai, Michael Terry, Quoc Le, et~al.
\newblock Program synthesis with large language models.
\newblock \emph{arXiv preprint arXiv:2108.07732}, 2021{\natexlab{b}}.

\bibitem[Bao et~al.(2024)Bao, Xiang, Yue, He, Zhu, Zheng, Zhao, Liu, Wang, and Zhu]{bao2024vidu}
Fan Bao, Chendong Xiang, Gang Yue, Guande He, Hongzhou Zhu, Kaiwen Zheng, Min Zhao, Shilong Liu, Yaole Wang, and Jun Zhu.
\newblock Vidu: a highly consistent, dynamic and skilled text-to-video generator with diffusion models.
\newblock \emph{arXiv preprint arXiv:2405.04233}, 2024.

\bibitem[Campbell et~al.(2022)Campbell, Benton, De~Bortoli, Rainforth, Deligiannidis, and Doucet]{campbell2022continuous}
Andrew Campbell, Joe Benton, Valentin De~Bortoli, Thomas Rainforth, George Deligiannidis, and Arnaud Doucet.
\newblock A continuous time framework for discrete denoising models.
\newblock \emph{Advances in Neural Information Processing Systems}, 35:\penalty0 28266--28279, 2022.

\bibitem[Chen et~al.(2024{\natexlab{a}})Chen, Mart{\'\i}~Mons{\'o}, Du, Simchowitz, Tedrake, and Sitzmann]{chen2024diffusion}
Boyuan Chen, Diego Mart{\'\i}~Mons{\'o}, Yilun Du, Max Simchowitz, Russ Tedrake, and Vincent Sitzmann.
\newblock Diffusion forcing: Next-token prediction meets full-sequence diffusion.
\newblock \emph{Advances in Neural Information Processing Systems}, 37:\penalty0 24081--24125, 2024{\natexlab{a}}.

\bibitem[Chen et~al.(2024{\natexlab{b}})Chen, Ge, Xie, Wu, Yao, Ren, Wang, Luo, Lu, and Li]{chen2024pixart}
Junsong Chen, Chongjian Ge, Enze Xie, Yue Wu, Lewei Yao, Xiaozhe Ren, Zhongdao Wang, Ping Luo, Huchuan Lu, and Zhenguo Li.
\newblock Pixart-$\sigma$: Weak-to-strong training of diffusion transformer for 4k text-to-image generation.
\newblock In \emph{European Conference on Computer Vision}, pp.\  74--91. Springer, 2024{\natexlab{b}}.

\bibitem[Chen et~al.(2021)Chen, Tworek, Jun, Yuan, de~Oliveira~Pinto, Kaplan, Edwards, Burda, Joseph, Brockman, Ray, Puri, Krueger, Petrov, Khlaaf, Sastry, Mishkin, Chan, Gray, Ryder, Pavlov, Power, Kaiser, Bavarian, Winter, Tillet, Such, Cummings, Plappert, Chantzis, Barnes, Herbert-Voss, Guss, Nichol, Paino, Tezak, Tang, Babuschkin, Balaji, Jain, Saunders, Hesse, Carr, Leike, Achiam, Misra, Morikawa, Radford, Knight, Brundage, Murati, Mayer, Welinder, McGrew, Amodei, McCandlish, Sutskever, and Zaremba]{chen2021codex}
Mark Chen, Jerry Tworek, Heewoo Jun, Qiming Yuan, Henrique~Ponde de~Oliveira~Pinto, Jared Kaplan, Harri Edwards, Yuri Burda, Nicholas Joseph, Greg Brockman, Alex Ray, Raul Puri, Gretchen Krueger, Michael Petrov, Heidy Khlaaf, Girish Sastry, Pamela Mishkin, Brooke Chan, Scott Gray, Nick Ryder, Mikhail Pavlov, Alethea Power, Lukasz Kaiser, Mohammad Bavarian, Clemens Winter, Philippe Tillet, Felipe~Petroski Such, Dave Cummings, Matthias Plappert, Fotios Chantzis, Elizabeth Barnes, Ariel Herbert-Voss, William~Hebgen Guss, Alex Nichol, Alex Paino, Nikolas Tezak, Jie Tang, Igor Babuschkin, Suchir Balaji, Shantanu Jain, William Saunders, Christopher Hesse, Andrew~N. Carr, Jan Leike, Josh Achiam, Vedant Misra, Evan Morikawa, Alec Radford, Matthew Knight, Miles Brundage, Mira Murati, Katie Mayer, Peter Welinder, Bob McGrew, Dario Amodei, Sam McCandlish, Ilya Sutskever, and Wojciech Zaremba.
\newblock Evaluating large language models trained on code.
\newblock 2021.

\bibitem[Cobbe et~al.(2021)Cobbe, Kosaraju, Bavarian, Chen, Jun, Kaiser, Plappert, Tworek, Hilton, Nakano, Hesse, and Schulman]{cobbe2021gsm8k}
Karl Cobbe, Vineet Kosaraju, Mohammad Bavarian, Mark Chen, Heewoo Jun, Lukasz Kaiser, Matthias Plappert, Jerry Tworek, Jacob Hilton, Reiichiro Nakano, Christopher Hesse, and John Schulman.
\newblock Training verifiers to solve math word problems.
\newblock \emph{arXiv preprint arXiv:2110.14168}, 2021.

\bibitem[Esser et~al.(2024)Esser, Kulal, Blattmann, Entezari, M{\"u}ller, Saini, Levi, Lorenz, Sauer, Boesel, et~al.]{esser2024scaling}
Patrick Esser, Sumith Kulal, Andreas Blattmann, Rahim Entezari, Jonas M{\"u}ller, Harry Saini, Yam Levi, Dominik Lorenz, Axel Sauer, Frederic Boesel, et~al.
\newblock Scaling rectified flow transformers for high-resolution image synthesis.
\newblock In \emph{Forty-first international conference on machine learning}, 2024.

\bibitem[{Google DeepMind}(2025)]{gemini2025}
{Google DeepMind}.
\newblock {Gemini Diffusion}.
\newblock \url{https://deepmind.google/models/gemini-diffusion}, 2025.
\newblock Accessed: 2025-05-24.

\bibitem[Grattafiori et~al.(2024)Grattafiori, Dubey, Jauhri, Pandey, Kadian, Al-Dahle, Letman, Mathur, Schelten, Vaughan, et~al.]{grattafiori2024llama}
Aaron Grattafiori, Abhimanyu Dubey, Abhinav Jauhri, Abhinav Pandey, Abhishek Kadian, Ahmad Al-Dahle, Aiesha Letman, Akhil Mathur, Alan Schelten, Alex Vaughan, et~al.
\newblock The llama 3 herd of models.
\newblock \emph{arXiv preprint arXiv:2407.21783}, 2024.

\bibitem[Guo et~al.(2025)Guo, Yang, Zhang, Song, Zhang, Xu, Zhu, Ma, Wang, Bi, et~al.]{guo2025deepseek}
Daya Guo, Dejian Yang, Haowei Zhang, Junxiao Song, Ruoyu Zhang, Runxin Xu, Qihao Zhu, Shirong Ma, Peiyi Wang, Xiao Bi, et~al.
\newblock Deepseek-r1: Incentivizing reasoning capability in llms via reinforcement learning.
\newblock \emph{arXiv preprint arXiv:2501.12948}, 2025.

\bibitem[Hendrycks et~al.(2021)Hendrycks, Burns, Kadavath, Arora, Basart, Tang, Song, and Steinhardt]{hendrycks2021measuring}
Dan Hendrycks, Collin Burns, Saurav Kadavath, Akul Arora, Steven Basart, Eric Tang, Dawn Song, and Jacob Steinhardt.
\newblock Measuring mathematical problem solving with the math dataset.
\newblock \emph{arXiv preprint arXiv:2103.03874}, 2021.

\bibitem[Hu et~al.(2022)Hu, Shen, Wallis, Allen-Zhu, Li, Wang, Wang, Chen, et~al.]{hu2022lora}
Edward~J Hu, Yelong Shen, Phillip Wallis, Zeyuan Allen-Zhu, Yuanzhi Li, Shean Wang, Lu~Wang, Weizhu Chen, et~al.
\newblock Lora: Low-rank adaptation of large language models.
\newblock \emph{ICLR}, 1\penalty0 (2):\penalty0 3, 2022.

\bibitem[Hu et~al.(2025)Hu, Meng, Akhauri, Abdelfattah, Seo, Zhang, and Gupta]{hu2025accelerating}
Zhanqiu Hu, Jian Meng, Yash Akhauri, Mohamed~S Abdelfattah, Jae-sun Seo, Zhiru Zhang, and Udit Gupta.
\newblock Accelerating diffusion language model inference via efficient kv caching and guided diffusion.
\newblock \emph{arXiv preprint arXiv:2505.21467}, 2025.

\bibitem[Huang et~al.(2025)Huang, Li, He, Zhou, and Shechtman]{huang2025self}
Xun Huang, Zhengqi Li, Guande He, Mingyuan Zhou, and Eli Shechtman.
\newblock Self forcing: Bridging the train-test gap in autoregressive video diffusion.
\newblock \emph{arXiv preprint arXiv:2506.08009}, 2025.

\bibitem[Inception et~al.(2025)Inception, Khanna, Kharbanda, Li, Varma, Wang, Birnbaum, Luo, Miraoui, Palrecha, et~al.]{labs2025mercury}
Labs Inception, Samar Khanna, Siddhant Kharbanda, Shufan Li, Harshit Varma, Eric Wang, Sawyer Birnbaum, Ziyang Luo, Yanis Miraoui, Akash Palrecha, et~al.
\newblock Mercury: Ultra-fast language models based on diffusion.
\newblock \emph{arXiv preprint arXiv:2506.17298}, 2025.

\bibitem[Israel et~al.(2025)Israel, Broeck, and Grover]{israel2025accelerating}
Daniel Israel, Guy Van~den Broeck, and Aditya Grover.
\newblock Accelerating diffusion llms via adaptive parallel decoding.
\newblock \emph{arXiv preprint arXiv:2506.00413}, 2025.

\bibitem[Labs(2025)]{bespoke_stratos}
Bespoke Labs.
\newblock Bespoke-stratos: The unreasonable effectiveness of reasoning distillation.
\newblock https://www.bespokelabs.ai/blog/bespoke-stratos-the-unreasonable-effectiveness-of-reasoning-distillation, 2025.
\newblock Accessed: 2025-01-22.

\bibitem[Labs et~al.(2025)Labs, Batifol, Blattmann, Boesel, Consul, Diagne, Dockhorn, English, English, Esser, Kulal, Lacey, Levi, Li, Lorenz, Müller, Podell, Rombach, Saini, Sauer, and Smith]{labs2025flux1kontextflowmatching}
Black~Forest Labs, Stephen Batifol, Andreas Blattmann, Frederic Boesel, Saksham Consul, Cyril Diagne, Tim Dockhorn, Jack English, Zion English, Patrick Esser, Sumith Kulal, Kyle Lacey, Yam Levi, Cheng Li, Dominik Lorenz, Jonas Müller, Dustin Podell, Robin Rombach, Harry Saini, Axel Sauer, and Luke Smith.
\newblock Flux.1 kontext: Flow matching for in-context image generation and editing in latent space, 2025.
\newblock URL \url{https://arxiv.org/abs/2506.15742}.

\bibitem[Liu et~al.(2024)Liu, Feng, Xue, Wang, Wu, Lu, Zhao, Deng, Zhang, Ruan, et~al.]{liu2024deepseek}
Aixin Liu, Bei Feng, Bing Xue, Bingxuan Wang, Bochao Wu, Chengda Lu, Chenggang Zhao, Chengqi Deng, Chenyu Zhang, Chong Ruan, et~al.
\newblock Deepseek-v3 technical report.
\newblock \emph{arXiv preprint arXiv:2412.19437}, 2024.

\bibitem[Liu et~al.(2025)Liu, Yang, Zhang, Chen, Zou, Wei, Wang, and Zhang]{liu2025dllm}
Zhiyuan Liu, Yicun Yang, Yaojie Zhang, Junjie Chen, Chang Zou, Qingyuan Wei, Shaobo Wang, and Linfeng Zhang.
\newblock dllm-cache: Accelerating diffusion large language models with adaptive caching.
\newblock \emph{arXiv preprint arXiv:2506.06295}, 2025.

\bibitem[Lou et~al.(2023)Lou, Meng, and Ermon]{lou2023discrete}
Aaron Lou, Chenlin Meng, and Stefano Ermon.
\newblock Discrete diffusion modeling by estimating the ratios of the data distribution.
\newblock \emph{arXiv preprint arXiv:2310.16834}, 2023.

\bibitem[Ma et~al.(2025)Ma, Yu, Fang, and Wang]{ma2025dkv}
Xinyin Ma, Runpeng Yu, Gongfan Fang, and Xinchao Wang.
\newblock dkv-cache: The cache for diffusion language models.
\newblock \emph{arXiv preprint arXiv:2505.15781}, 2025.

\bibitem[Nie et~al.(2025)Nie, Zhu, You, Zhang, Ou, Hu, Zhou, Lin, Wen, and Li]{nie2025large}
Shen Nie, Fengqi Zhu, Zebin You, Xiaolu Zhang, Jingyang Ou, Jun Hu, Jun Zhou, Yankai Lin, Ji-Rong Wen, and Chongxuan Li.
\newblock Large language diffusion models.
\newblock \emph{arXiv preprint arXiv:2502.09992}, 2025.

\bibitem[Peng et~al.(2025)Peng, Zheng, Shen, Young, Guo, Wang, Xu, Liu, Jiang, Li, et~al.]{peng2025open}
Xiangyu Peng, Zangwei Zheng, Chenhui Shen, Tom Young, Xinying Guo, Binluo Wang, Hang Xu, Hongxin Liu, Mingyan Jiang, Wenjun Li, et~al.
\newblock Open-sora 2.0: Training a commercial-level video generation model in $200$ k.
\newblock \emph{arXiv preprint arXiv:2503.09642}, 2025.

\bibitem[Po et~al.(2025)Po, Nitzan, Zhang, Chen, Dao, Shechtman, Wetzstein, and Huang]{po2025long}
Ryan Po, Yotam Nitzan, Richard Zhang, Berlin Chen, Tri Dao, Eli Shechtman, Gordon Wetzstein, and Xun Huang.
\newblock Long-context state-space video world models.
\newblock \emph{arXiv preprint arXiv:2505.20171}, 2025.

\bibitem[Podell et~al.(2023)Podell, English, Lacey, Blattmann, Dockhorn, M{\"u}ller, Penna, and Rombach]{podell2023sdxl}
Dustin Podell, Zion English, Kyle Lacey, Andreas Blattmann, Tim Dockhorn, Jonas M{\"u}ller, Joe Penna, and Robin Rombach.
\newblock Sdxl: Improving latent diffusion models for high-resolution image synthesis.
\newblock \emph{arXiv preprint arXiv:2307.01952}, 2023.

\bibitem[Song \& Zhou(2025)Song and Zhou]{song2025ideas}
Jiaming Song and Linqi Zhou.
\newblock Ideas in inference-time scaling can benefit generative pre-training algorithms.
\newblock \emph{arXiv preprint arXiv:2503.07154}, 2025.

\bibitem[Sun et~al.(2025)Sun, Wang, Li, Liu, Sun, Feng, Lao, Zhou, He, and Liu]{sun2025ar}
Mingzhen Sun, Weining Wang, Gen Li, Jiawei Liu, Jiahui Sun, Wanquan Feng, Shanshan Lao, SiYu Zhou, Qian He, and Jing Liu.
\newblock Ar-diffusion: Asynchronous video generation with auto-regressive diffusion.
\newblock In \emph{Proceedings of the Computer Vision and Pattern Recognition Conference}, pp.\  7364--7373, 2025.

\bibitem[Teng et~al.(2025)Teng, Jia, Sun, Li, Li, Tang, Han, Zhang, Zhang, Luo, et~al.]{teng2025magi}
Hansi Teng, Hongyu Jia, Lei Sun, Lingzhi Li, Maolin Li, Mingqiu Tang, Shuai Han, Tianning Zhang, WQ~Zhang, Weifeng Luo, et~al.
\newblock Magi-1: Autoregressive video generation at scale.
\newblock \emph{arXiv preprint arXiv:2505.13211}, 2025.

\bibitem[Touvron et~al.(2023{\natexlab{a}})Touvron, Lavril, Izacard, Martinet, Lachaux, Lacroix, Rozi{\`e}re, Goyal, Hambro, Azhar, et~al.]{touvron2023llama}
Hugo Touvron, Thibaut Lavril, Gautier Izacard, Xavier Martinet, Marie-Anne Lachaux, Timoth{\'e}e Lacroix, Baptiste Rozi{\`e}re, Naman Goyal, Eric Hambro, Faisal Azhar, et~al.
\newblock Llama: Open and efficient foundation language models.
\newblock \emph{arXiv preprint arXiv:2302.13971}, 2023{\natexlab{a}}.

\bibitem[Touvron et~al.(2023{\natexlab{b}})Touvron, Martin, Stone, Albert, Almahairi, Babaei, Bashlykov, Batra, Bhargava, Bhosale, et~al.]{touvron2023llama2}
Hugo Touvron, Louis Martin, Kevin Stone, Peter Albert, Amjad Almahairi, Yasmine Babaei, Nikolay Bashlykov, Soumya Batra, Prajjwal Bhargava, Shruti Bhosale, et~al.
\newblock Llama 2: Open foundation and fine-tuned chat models.
\newblock \emph{arXiv preprint arXiv:2307.09288}, 2023{\natexlab{b}}.

\bibitem[Wei et~al.(2025)Wei, Zhang, Liu, Liu, and Zhang]{wei2025accelerating}
Qingyan Wei, Yaojie Zhang, Zhiyuan Liu, Dongrui Liu, and Linfeng Zhang.
\newblock Accelerating diffusion large language models with slowfast: The three golden principles.
\newblock \emph{arXiv preprint arXiv:2506.10848}, 2025.

\bibitem[Wu et~al.(2025)Wu, Zhang, Xue, Liu, Diao, Zhu, Luo, Han, and Xie]{wu2025fast}
Chengyue Wu, Hao Zhang, Shuchen Xue, Zhijian Liu, Shizhe Diao, Ligeng Zhu, Ping Luo, Song Han, and Enze Xie.
\newblock Fast-dllm: Training-free acceleration of diffusion llm by enabling kv cache and parallel decoding.
\newblock \emph{arXiv preprint arXiv:2505.22618}, 2025.

\bibitem[Yang et~al.(2024{\natexlab{a}})Yang, Yang, Zhang, Hui, Zheng, Yu, Li, Liu, Huang, Wei, Lin, Yang, Tu, Zhang, Yang, Yang, Zhou, Lin, Dang, Lu, Bao, Yang, Yu, Li, Xue, Zhang, Zhu, Men, Lin, Li, Xia, Ren, Ren, Fan, Su, Zhang, Wan, Liu, Cui, Zhang, and Qiu]{qwen2.5}
An~Yang, Baosong Yang, Beichen Zhang, Binyuan Hui, Bo~Zheng, Bowen Yu, Chengyuan Li, Dayiheng Liu, Fei Huang, Haoran Wei, Huan Lin, Jian Yang, Jianhong Tu, Jianwei Zhang, Jianxin Yang, Jiaxi Yang, Jingren Zhou, Junyang Lin, Kai Dang, Keming Lu, Keqin Bao, Kexin Yang, Le~Yu, Mei Li, Mingfeng Xue, Pei Zhang, Qin Zhu, Rui Men, Runji Lin, Tianhao Li, Tingyu Xia, Xingzhang Ren, Xuancheng Ren, Yang Fan, Yang Su, Yichang Zhang, Yu~Wan, Yuqiong Liu, Zeyu Cui, Zhenru Zhang, and Zihan Qiu.
\newblock Qwen2.5 technical report.
\newblock \emph{arXiv preprint arXiv:2412.15115}, 2024{\natexlab{a}}.

\bibitem[Yang et~al.(2025)Yang, Li, Yang, Zhang, Hui, Zheng, Yu, Gao, Huang, Lv, et~al.]{yang2025qwen3}
An~Yang, Anfeng Li, Baosong Yang, Beichen Zhang, Binyuan Hui, Bo~Zheng, Bowen Yu, Chang Gao, Chengen Huang, Chenxu Lv, et~al.
\newblock Qwen3 technical report.
\newblock \emph{arXiv preprint arXiv:2505.09388}, 2025.

\bibitem[Yang et~al.(2024{\natexlab{b}})Yang, Teng, Zheng, Ding, Huang, Xu, Yang, Hong, Zhang, Feng, et~al.]{yang2024cogvideox}
Zhuoyi Yang, Jiayan Teng, Wendi Zheng, Ming Ding, Shiyu Huang, Jiazheng Xu, Yuanming Yang, Wenyi Hong, Xiaohan Zhang, Guanyu Feng, et~al.
\newblock Cogvideox: Text-to-video diffusion models with an expert transformer.
\newblock \emph{arXiv preprint arXiv:2408.06072}, 2024{\natexlab{b}}.

\bibitem[Ye et~al.(2025)Ye, Xie, Zheng, Gao, Wu, Jiang, Li, and Kong]{dream2025}
Jiacheng Ye, Zhihui Xie, Lin Zheng, Jiahui Gao, Zirui Wu, Xin Jiang, Zhenguo Li, and Lingpeng Kong.
\newblock Dream 7b, 2025.
\newblock URL \url{https://hkunlp.github.io/blog/2025/dream}.

\bibitem[Yin et~al.(2025)Yin, Zhang, Zhang, Freeman, Durand, Shechtman, and Huang]{yin2025slow}
Tianwei Yin, Qiang Zhang, Richard Zhang, William~T Freeman, Fredo Durand, Eli Shechtman, and Xun Huang.
\newblock From slow bidirectional to fast autoregressive video diffusion models.
\newblock In \emph{Proceedings of the Computer Vision and Pattern Recognition Conference}, pp.\  22963--22974, 2025.

\bibitem[Zheng et~al.(2024)Zheng, Peng, Yang, Shen, Li, Liu, Zhou, Li, and You]{zheng2024open}
Zangwei Zheng, Xiangyu Peng, Tianji Yang, Chenhui Shen, Shenggui Li, Hongxin Liu, Yukun Zhou, Tianyi Li, and Yang You.
\newblock Open-sora: Democratizing efficient video production for all.
\newblock \emph{arXiv preprint arXiv:2412.20404}, 2024.

\bibitem[Zhu et~al.(2025)Zhu, Wang, Nie, Zhang, Wu, Hu, Zhou, Chen, Lin, Wen, et~al.]{zhu2025llada}
Fengqi Zhu, Rongzhen Wang, Shen Nie, Xiaolu Zhang, Chunwei Wu, Jun Hu, Jun Zhou, Jianfei Chen, Yankai Lin, Ji-Rong Wen, et~al.
\newblock Llada 1.5: Variance-reduced preference optimization for large language diffusion models.
\newblock \emph{arXiv preprint arXiv:2505.19223}, 2025.

\end{thebibliography}
\bibliographystyle{iclr2025_conference}

\newpage
\appendix

\section{Appendix}

\label{sec:appendix}

\subsection{Ablation Studies on D2F Components}
To isolate the performance gains from different components of our D2F framework, we conduct ablation studies on both the \textbf{D2F-LLaDA} (Table~\ref{tab:ablation_d2f_llada}) and \textbf{D2F-Dream} (Table~\ref{tab:ablation_d2f_dream}) models. We compare two configurations:
\begin{itemize}
    \item \textbf{Cache-only}: This configuration utilizes the block-wise causal attention mechanism to enable standard KV cache but does not employ the parallel, asynchronous block generation pipeline. Generation proceeds serially, one block at a time.
    \item \textbf{Cache + Para}: This is the full D2F method, combining KV cache with our pipelined parallel decoding algorithm.
\end{itemize}
The results clearly demonstrate that while enabling KV cache (Cache-only) provides a substantial speedup, the addition of our parallel decoding pipeline (Cache + Para) further accelerates inference by a significant margin (e.g., from 2.4x to 7.3x on GSM8K for LLaDA), highlighting the efficacy of the asynchronous generation strategy.

\subsection{Control Experiment for Data Contribution}
\label{app:control_exp}
To prove that our performance gains stem from the D2F framework and not the training data, we conducted a control experiment. We created a control model, Dream-Base*, by directly fine-tuning the original Dream-Base on the same distillation dataset. For a fair comparison, this fine-tuning used LoRA with the exact same parameters as our D2F training.

The results are in Table~\ref{tab:dream_base_vs_finetuned}. While Dream-Base* shows minor score improvements on some tasks, its inference speed drops significantly. This speed degradation is due to the computational overhead of the added LoRA layers. This control experiment confirms that our D2F methodology, rather than the data, is the key driver of the exceptional inference acceleration.

\subsection{Hyperparameter Details}
\label{sec:hyperparameter}
Table~\ref{tab:hyperparams} details the specific hyperparameter configurations used for both the baseline models and our D2F models across all evaluated benchmarks. Baseline settings are adopted from their respective original works to ensure fair comparison. For our D2F models, we specify the maximum generation length, the inference block size, and the key thresholds for our pipelined parallel decoding algorithm: the token addition threshold ($\tau_{add}$), the block activation threshold ($\tau_{act}$), and the token confirmation confidence ($\tau_{conf}$).

\begin{table*}[!htbp]
    \centering
    
    \definecolor{brightgreen}{rgb}{0.1, 0.65, 0.2}
    \newcommand{\speedup}[1]{\textcolor{brightgreen}{(#1)}}
    
    \setlength{\tabcolsep}{3pt}
    
    \sisetup{detect-weight=true, detect-family=true}
    
    \begin{tabular}{
        l  
        l  
        l  
        l  
        S[table-format=3.0] 
        S[table-format=2.1] 
    }
        \toprule
        \textbf{Test Set} & \textbf{Configuration} & \textbf{TPS ↑} & \textbf{Latency (s) ↓} & {\textbf{Gen. Length}} & {\textbf{Score ↑}} \\
        \midrule
        
        \multirow{2}{*}{\makecell[l]{\textbf{GSM8K} \\ \small 4-shot}}
        & Cache-only         & 17.5 \speedup{2.4x}  & 8.3 \speedup{3.9x}  & 145 & \textbf{78.1} \\
        & Cache + Para & \textbf{52.5} \speedup{7.3x}  & \textbf{2.8} \speedup{11.5x} & 144 & 77.3          \\
        \midrule
        
        \multirow{2}{*}{\makecell[l]{\textbf{MBPP} \\ \small 3-shot}}
        & Cache-only         & 18.1 \speedup{20.1x} & 3.8 \speedup{18.8x} & 69  & 37.6          \\
        & Cache + Para & \textbf{47.6} \speedup{52.9x} & \textbf{1.4} \speedup{51.0x} & 68  & \textbf{38.0} \\
        \midrule
        
        \multirow{2}{*}{\makecell[l]{\textbf{HumanEval} \\ \small 0-shot}}
        & Cache-only         & 28.0 \speedup{10.0x} & 4.8 \speedup{8.1x}  & 135 & \textbf{40.2} \\
        & Cache + Para & \textbf{81.6} \speedup{29.1x} & \textbf{1.6} \speedup{24.3x} & 133 & \textbf{40.2} \\
        \midrule
        
        \multirow{2}{*}{\makecell[l]{\textbf{Math} \\ \small 4-shot}}
        & Cache-only         & 30.6 \speedup{1.5x}  & 12.6 \speedup{0.9x} & 385 & \textbf{29.6} \\
        & Cache + Para & \textbf{90.2} \speedup{4.3x}  & \textbf{4.3} \speedup{2.7x}  & 384 & 29.1          \\
        \bottomrule
    \end{tabular}
    
    \caption{
        Ablation study of our proposed method (\textbf{D2F-LLaDA}) on the \textbf{LLaDA-Instruct} model. 
        Performance ratios relative to the LLaDA-Instruct baseline (from Table~\ref{tab:performance_comparison_final_spaced}) are shown in \speedup{bright green}.
    }
    \label{tab:ablation_d2f_llada}
\end{table*}

\begin{table*}[!htbp]
    \centering
    \definecolor{brightgreen}{rgb}{0.1, 0.65, 0.2}
    \newcommand{\speedup}[1]{\textcolor{brightgreen}{(#1)}}
    
    \setlength{\tabcolsep}{8pt}
    
    \sisetup{detect-weight=true, detect-family=true}
    
    \begin{tabular}{
        l  
        l  
        l  
        l  
        S[table-format=3.0] 
        S[table-format=2.1] 
    }
        \toprule
        \textbf{Test Set} & \textbf{Method} & \textbf{TPS ↑} & \textbf{Latency (s) ↓} & {\textbf{Gen. Length}} & {\textbf{Score ↑}} \\
        \midrule
        
        \multirow{2}{*}{\makecell[l]{\textbf{GSM8K-CoT} \\ \small 8-shot}}
        & Dream-Base    & \textbf{9.5} \speedup{1.0x}  & \textbf{26.8} \speedup{1.0x} & 255 & 75.0          \\
        & Dream-Base*   & 6.7 \speedup{0.7x}  & 29.9 \speedup{0.9x} & 199 & \textbf{77.8} \\
        \midrule
        
        \multirow{2}{*}{\makecell[l]{\textbf{MBPP} \\ \small 3-shot}}
        & Dream-Base    & \textbf{10.4} \speedup{1.0x} & \textbf{24.6} \speedup{1.0x} & 256 & 56.2          \\
        & Dream-Base*   & 4.2 \speedup{0.4x}  & 27.4 \speedup{0.9x} & 114 & \textbf{57.4} \\
        \midrule
        
        \multirow{2}{*}{\makecell[l]{\textbf{HumanEval} \\ \small 0-shot}}
        & Dream-Base    & \textbf{20.2} \speedup{1.0x} & \textbf{12.6} \speedup{1.0x} & 255 & \textbf{54.3} \\
        & Dream-Base*   & 8.8 \speedup{0.4x}  & 14.2 \speedup{0.9x} & 125 & 51.8          \\
        \midrule
        
        \multirow{2}{*}{\makecell[l]{\textbf{Math} \\ \small 4-shot}}
        & Dream-Base    & \textbf{9.9} \speedup{1.0x}  & \textbf{25.8} \speedup{1.0x} & 256 & \textbf{35.8} \\
        & Dream-Base*   & 5.4 \speedup{0.5x}  & 28.6 \speedup{0.9x} & 154 & 33.4          \\
        \bottomrule
    \end{tabular}
    
    \caption{
        Comparison between the baseline Dream-Base model and its fine-tuned version, Dream-Base*. 
        The fine-tuned version, highlighted with a gray background, is obtained by directly fine-tuning on the distillation data. 
        Performance ratios relative to the baseline are shown in \speedup{bright green}, and the highest score in each test set is marked in \textbf{bold}.
    }
    \label{tab:dream_base_vs_finetuned}
\end{table*}

\begin{table*}[!htbp]
    \centering
    
    \definecolor{brightgreen}{rgb}{0.1, 0.65, 0.2}
    \newcommand{\speedup}[1]{\textcolor{brightgreen}{(#1)}}
    
    \setlength{\tabcolsep}{3pt}
    
    \sisetup{detect-weight=true, detect-family=true}
    
    \begin{tabular}{
        l  
        l  
        l  
        l  
        S[table-format=3.0] 
        S[table-format=2.1] 
    }
        \toprule
        \textbf{Test Set} & \textbf{Configuration} & \textbf{TPS ↑} & \textbf{Latency (s) ↓} & {\textbf{Gen. Length}} & {\textbf{Score ↑}} \\
        \midrule
        
        \multirow{2}{*}{\makecell[l]{\textbf{GSM8K-CoT} \\ \small 8-shot}}
        & Cache-only         & 40.7 \speedup{4.3x}  & 6.3 \speedup{4.3x}  & 256 & \textbf{77.8} \\
        & Cache + Para & \textbf{91.2} \speedup{9.6x}  & \textbf{2.8} \speedup{9.6x}  & 256 & 77.6          \\
        \midrule
        
        \multirow{2}{*}{\makecell[l]{\textbf{MBPP} \\ \small 3-shot}}
        & Cache-only         & 40.5 \speedup{3.9x}  & 5.9 \speedup{4.2x}  & 240 & 54.2          \\
        & Cache + Para & \textbf{105} \speedup{10.1x} & \textbf{2.3} \speedup{10.7x} & 240 & \textbf{55.2} \\
        \midrule
        
        \multirow{2}{*}{\makecell[l]{\textbf{HumanEval} \\ \small 0-shot}}
        & Cache-only         & 43.6 \speedup{2.2x}  & 5.1 \speedup{2.5x}  & 222 & 53.7          \\
        & Cache + Para & \textbf{73.2} \speedup{3.6x}  & \textbf{3.1} \speedup{4.1x}  & 227 & \textbf{54.3} \\
        \midrule
        
        \multirow{2}{*}{\makecell[l]{\textbf{Math} \\ \small 4-shot}}
        & Cache-only         & 39.6 \speedup{4.0x}  & 6.5 \speedup{4.0x}  & 256 & \textbf{35.9} \\
        & Cache + Para & \textbf{98.8} \speedup{10.0x} & \textbf{2.6} \speedup{9.9x}  & 256 & 35.4          \\
        \bottomrule
    \end{tabular}
    
    \caption{
        Ablation study of our proposed method (\textbf{D2F-Dream}) on the \textbf{Dream-Base} model. 
        Performance ratios relative to the Dream-Base baseline (from Table~\ref{tab:acceleration_methods_comparison_optimized}) are shown in \speedup{bright green}.
    }
    \label{tab:ablation_d2f_dream} 
\end{table*}

\begin{table}[htbp]
\centering
\begin{tabular}{@{}ll ccccc@{}}
\toprule
\textbf{Benchmark} & \textbf{Configuration} & \textbf{length} & \textbf{block\_size} & $\boldsymbol{\tau_{add}}$ & $\boldsymbol{\tau_{act}}$ & $\boldsymbol{\tau_{conf}}$ \\
\midrule
\multicolumn{7}{@{}l}{\textbf{Method: D2F-LLaDA}} \\
\addlinespace[0.3em]
\multirow{2}{*}{GSM8K} & Baseline & 256 & 8 & -- & -- & -- \\
 & D2F & 512 & 64 & 0.7 & 0.95 & 0.9 \\
\cmidrule(l){2-7}
\multirow{2}{*}{MBPP} & Baseline & 512 & 32 & -- & -- & -- \\
 & D2F & 512 & 32 & 0.9 & 0.95 & 0.9 \\
\cmidrule(l){2-7}
\multirow{2}{*}{HumanEval} & Baseline & 512 & 32 & -- & -- & -- \\
 & D2F & 512 & 32 & 0.1 & 0.95 & 0.9 \\
\cmidrule(l){2-7}
\multirow{2}{*}{MATH} & Baseline & 256 & 256 & -- & -- & -- \\
 & D2F & 512 & 32 & 0.1 & 0.95 & 0.9 \\
\midrule
\multicolumn{7}{@{}l}{\textbf{Method: D2F-Dream}} \\
\addlinespace[0.3em]
\multirow{2}{*}{GSM8K-COT} & Baseline & 256 & -- & -- & -- & -- \\
 & D2F & 256 & 32 & 0.1 & 0.95 & 0.9 \\
\cmidrule(l){2-7}
\multirow{2}{*}{MBPP} & Baseline & 256 & -- & -- & -- & -- \\
 & D2F & 256 & 48 & 0.1 & 0.95 & 0.9 \\
\cmidrule(l){2-7}
\multirow{2}{*}{HumanEval} & Baseline & 256 & -- & -- & -- & -- \\
 & D2F & 256 & 32 & 0.9 & 0.95 & 0.95 \\
\cmidrule(l){2-7}
\multirow{2}{*}{MATH} & Baseline & 256 & -- & -- & -- & -- \\
 & D2F & 256 & 64 & 0.1 & 0.95 & 0.9 \\
\bottomrule
\end{tabular}
\caption{Sampling hyperparameters for the experiments in Table~\ref{tab:performance_comparison_final_spaced} and Table~\ref{tab:acceleration_methods_comparison_optimized}. Baseline settings are adopted from prior work. For D2F, the \texttt{length} parameter only sets a maximum generation limit without affecting the sampling distribution due to its block-wise nature. Notably, for the Dream-based model, we match D2F's maximum length to the baseline's, as the base model often fails to generate proper termination tokens.}
\label{tab:hyperparams}
\end{table}


\end{document}